\documentclass[journal,12pt,draftclsnofoot,onecolumn]{IEEEtran}

\ifCLASSINFOpdf
\else
\fi
\usepackage{hyperref}
\usepackage{url}
\usepackage{subfigure}%
\usepackage{bm}              %
\usepackage{epsfig}
\usepackage[flushmargin]{footmisc}
\usepackage{float}
\usepackage{multirow}
\begin{document}
%

\title{\quad \\ Image Set based Collaborative Representation for Face Recognition}

\author{
{\quad \\
\normalsize
        Pengfei Zhu,~\IEEEmembership{\normalsize Student Member,~IEEE,}
        Wangmeng Zuo, ~\IEEEmembership{\normalsize Member,~IEEE,}
        Lei Zhang, ~\IEEEmembership{\normalsize Member,~IEEE,}
        Simon C.K. Shiu, ~\IEEEmembership{\normalsize Member,~IEEE,}
        David Zhang,  ~\IEEEmembership{\normalsize Fellow,~IEEE}
\IEEEcompsocitemizethanks{The authors are with Department of Computing
The Hong Kong Polytechnic University Hung Hom, Kowloon, Hong Kong, China (e-mail: cspzhu, cswmzuo, cslzhang, csckshiu, csdzhang@comp.polyu.edu.hk).\protect\\
}
}
}


\maketitle

\begin{abstract}
With the rapid development of digital imaging and communication technologies, image set based face recognition (ISFR) is becoming increasingly important. One key issue of ISFR is how to effectively and efficiently represent the query face image set by using the gallery face image sets. The set-to-set distance based methods ignore the relationship between gallery sets, while representing the query set images individually over the gallery sets ignores the correlation between query set images. In this paper, we propose a novel image set based collaborative representation and classification method for ISFR. By modeling the query set as a convex or regularized hull, we represent this hull collaboratively over all the gallery sets. With the resolved representation coefficients, the distance between the query set and each gallery set can then be calculated for classification. The proposed model naturally and effectively extends the image based collaborative representation to an image set based one, and our extensive experiments on benchmark ISFR databases show the superiority of the proposed method to state-of-the-art ISFR methods under different set sizes in terms of both recognition rate and efficiency.
\end{abstract}


\newpage
\section{Introduction}

Image set based classification has been increasingly employed in face recognition \cite{yamaguchi1998face,arandjelovic2005face,nishiyama2007recognizing,wang2008manifold,cevikalp2010face,wolf2011face,hu2011sparse,cui2012image,chen2012dictionary} and object categorization \cite{kim2007discriminative,wang2009manifold} in recent years. Due to the rapid development of digital imaging and communication techniques, now image sets can be easily collected from multi-view images using multiple cameras \cite{kim2007discriminative}, long term observations \cite{wolf2011face}, personal albums and news pictures \cite{LFWTech}, etc. Meanwhile, image set based face recognition (ISFR) has shown superior performance to single image based face recognition since the many sample images in the gallery set can convey more within-class variations of the subject \cite{hu2011sparse}. One special case of ISFR is video based face recognition, which collects face image sets from consecutive video sequences \cite{yamaguchi1998face,lee2003video,stallkamp2007video}. Similar to the work in \cite{cevikalp2010face,{hu2011sparse}}, in this paper we focus on the general case of ISFR without considering the temporal relationship of samples in each set.

The key issues in image set based classification include how to model a set and consequently how to compute the distance/similarity between query and gallery sets. Researchers have proposed parametric and non-parametric approaches for image set modeling. Parametric modeling methods model each set as a parametric distribution, and use Kullback-Leibler divergence to measure the similarity between the distributions \cite{arandjelovic2005face, wolf2011face}. The disadvantage of parametric set modeling lies in the difficulty of parameter estimation, and it may fail when the estimated parametric model does not fit well the real gallery and query sets \cite{kim2007discriminative,wang2008manifold,hu2011sparse}.

Many non-parametric set modeling methods have also been proposed, including subspace \cite{kim2007discriminative,yamaguchi1998face,TIPCMSM},
manifold \cite{hadid2004still,fan2006locally,wang2008manifold,wang2009manifold,wang2012covariance},
affine hull \cite{cevikalp2010face,hu2011sparse},
convex hull \cite{cevikalp2010face},
and covariance matrix based ones \cite{wang2012covariance, caseirorolling, jayasumanakernel}.
The method in \cite{kim2007discriminative} employs canonical correlation to measure the similarity between two sets.
A projection matrix is learned by maximizing the canonical correlations of within-class sets while
minimizing the canonical correlations of between-class sets.
The methods in \cite{TIPMMD} use manifold to model an image set and define a manifold-to-manifold distance (MMD) for set matching.
MMD models each image set as a set of local subspaces and the distance between two image sets is defined as
a weighted average of pairwise subspace to subspace distance.
As MMD is a non-discriminative measure, Manifold Discriminant Analysis (MDA) is proposed to learn an embedding space by maximizing manifold margin \cite{wang2009manifold}.
The performance of subspace and manifold based methods may degrade much when the set has a small sample size but big data variations \cite{hu2011sparse,wang2012covariance}.
In affine hull and convex hull based methods \cite{cevikalp2010face,hu2011sparse}, the between-set distance is defined as the distance between the two closest points of the two sets.
When convex hull is used, the set to set distance is equivalent to the nearest point problem in SVM \cite{burges2000geometric}.
In \cite{hu2012face}, a method called sparse approximated nearest points (SANP) is proposed to measure the dissimilarity between two image sets.
To reduce the model complexity of SANP, a reduced model, which is called regularized nearest points (RNP), is proposed by modeling each image set as a regularized hull \cite{yang2013face}.
However, the closest points based methods \cite{cevikalp2010face,hu2011sparse,wu2012set,{yang2013face}} rely highly on the location of each individual sample in the set, and the model fitting can be heavily deteriorated by outliers \cite{wang2012covariance}.
In \cite{wang2012covariance}, an image set is represented by a covariance matrix and a Riemannian kernel function is defined to measure the similarity between two image sets by a mapping from the Riemannian manifold to a Euclidean space.
With the kernel function between two image sets, traditional discriminant learning methods, e.g., linear discriminative analysis \cite{baudat2000generalized}, partial least squares \cite{rosipal2006overview}, kernel machines, can be used for image set classification \cite{caseirorolling, jayasumanakernel}.
The disadvantages of covariance matrix based methods include the computational complexity of eigen-decomposition of symmetric positive-definite (SPD) matrices and the curse of dimensionality with limited number of training sets.

No matter how the set is modeled, in almost all the previous works \cite{kim2007discriminative,yamaguchi1998face,hadid2004still,fan2006locally,wang2008manifold,wang2009manifold,wang2012covariance,cevikalp2010face,hu2011sparse,{yang2013face}}, the query set is compared to each of the gallery sets separately, and then classified to the class closest to it. Such a classification scheme does not consider the correlation between gallery sets, like the nearest neighbor or nearest subspace classifier in single image based face recognition. In recent years, the sparse representation based classification (SRC) \cite{wright2009robust} has shown interesting results in image based face recognition. SRC represents a query face as a sparse linear combination of samples from all classes, and classifies it to the class which has the minimal representation residual to it.
Though SRC emphasizes much on the role of $l_1$-norm sparsity of representation coefficients, it has been shown in \cite{zhang2011sparse} that the collaborative representation mechanism (i.e., using samples from all classes to collaboratively represent the query image) is more important to the success of SRC.
The so-called collaborative representation based classification (CRC) with $l_2$-regularization leads to similar results to SRC but with much lower computational cost \cite{zhang2011sparse}.
In \cite{TIPRSC}, feature weights are introduced to the representation model to penalize pixels with large error so that the model is robust to outliers.
Moreover, a kernel sparse representation model is proposed for face recognition by mapping features to a high dimensional Reproducing Kernel Hilbert Space (RKHS), which further improves the recognition accuracy \cite{gao2010kernel,gao2013sparse}.
Similarly, a robust kernel representation model is proposed with iteratively reweighted algorithms \cite{yang2013}.

One may apply SRC/CRC to ISFR by representing each image of the query set over all the gallery sets, and then using the average or minimal representation residual of the query set images for classification. However, such a scheme does not exploit the correlation and distinctiveness of sample images in the query set. If the average representation residual is used for classification, the discrimination of representation residuals by different classes will be reduced; if the minimal representation residual is used, the classification can suffer from the outlier images in the query set.
In addition, there are redundancies in an image set.
The redundancies will lead to great storage burden and computational complexity, and deteriorate the recognition performance.

\begin{figure}[h]
\centering
\includegraphics[width=12cm]{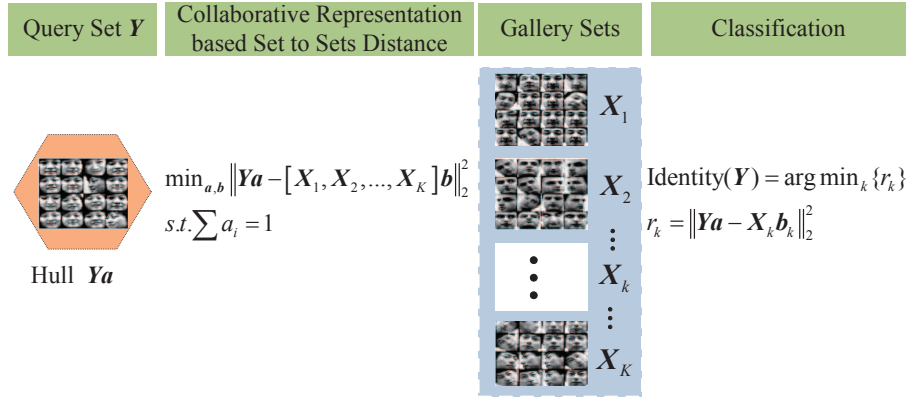}
\caption{Image set based collaborative representation and classification (ISCRC).}
\label{flowcart}
\end{figure}

In this paper, we propose a novel image set based collaborative representation and classification (ISCRC) approach for ISFR, as illustrated in Fig. \ref{flowcart}. The query set, denoted by $\bm{Y}$ (each column of $\bm{Y}$ is an image in the set) is modeled as a hull ${\bm{Y}}{\bm{a}}$ with the sum of coefficients in $\bm{a}$ being 1. Let ${\bm{X}}_k, k=1,2,...,K,$ be a gallery set. We then propose a collaborative representation based set (i.e., $\bm{Y}$) to sets (i.e., ${\bm{X}} = [{{\bm{X}}_1},...,{{\bm{X}}_k},...,{{\bm{X}}_K}]$) distance (CRSSD for short); that is, we represent the hull ${\bm{Y}}{\bm{a}}$ over the gallery sets $\bm{X}$ as ${\bm{X}}{\bm{b}}$, where $\bm{b}$ is a coefficient vector. Consequently, we can classify the query set $\bm{Y}$ by checking which gallery set has the minimal representation residual to the hull ${\bm{Y}}{\bm{a}}$. To get a stable solution to CRSSD, regularizations can be imposed on $\bm{a}$ and $\bm{b}$. In the proposed ISCRC, the gallery sets ${\bm{X}}_k$ can be compressed to a smaller size to remove the redundancy so that the time complexity of ISCRC can be much reduced without sacrificing the recognition rate. Our experiments on three benchmark ISFR databases show that the proposed ISCRC is superior to state-of-the-art methods in terms of both recognition rate and efficiency.

This paper is organized as follows. Section \ref{section2} discusses in detail the proposed CRSSD and ISCRC methods. Section \ref{section3} presents the regularized hull based ISCRC, followed by the convex hull based ISCRC in Section \ref{section4}. Section \ref{section5} conducts experiments and Section \ref{section6} gives our conclusions.
The main abbreviations used in the development of our method are summarized in Table \ref{abbreviations}.
\renewcommand{\multirowsetup}{\centering}
\begin{table}[h] \small
\begin{center}
\caption{The main abbreviations used in this paper}
\tabcolsep=0.2in
\begin{tabular}{c|c}
\hline
ISFR &  image set based face recognition\\
\hline
SRC &  sparse representation based classification\\
\hline
CRC &  collaborative representation based classification\\
\hline
\multirow{2}{1 cm}{CRSSD}
& collaborative representation based\\
& set to sets distance\\
\hline
\multirow{2}{1 cm}{ISCRC}
& image set based collaborative\\
& representation and classification\\
\hline
RH-ISCRC & regularized hull based ISCRC\\
\hline
KCH-ISCRC & kernelized convex hull based ISCRC\\
\hline
\end{tabular}
\label{abbreviations}
\end{center}
\end{table}

\section{Collaborative representation based set to sets distance}
\label{section2}
We first introduce the hull based set to set distance in \ref{section21}, and then propose the collaborative representation based set to sets distance (CRSSD) in \ref{section22}. With CRSSD, the image set based collaborative representation and classification (ISCRC) scheme can be naturally proposed. In \ref{section23} and \ref{section24}, the convex hull and regularized hull based CRSSD are respectively presented.

\subsection{Hull based set to set distance}
\label{section21}

In image set based classification, compared to the parametric modeling of image set, non-parametric modeling does not impose assumptions on the data distribution and inherits many favorable properties \cite{kim2007discriminative,hu2011sparse,wang2012covariance}. One simple non-parametric set modeling approach is the hull based modeling \cite{cevikalp2010face,hu2011sparse}, which models a set as the linear combination of its samples. Given a sample set $\bm{Y} = \{ {\bm{y}_1},...,{\bm{y}_i},...,{\bm{y}_n}\}$, ${\bm{y}_i} \in {{\Re}^d}$, the hull of set $\bm{Y}$ is defined as: $H(\bm{Y}) = \{ \sum {{a_i}{\bm{y}_i}}\}$.
Usually, $\sum\limits {{a_i}} = 1$ is required and the coefficients $a_i$ are required to be bounded:
\begin{eqnarray}
 \begin{array}{l}
H({\bm{Y}}) = \left\{ {\sum {{a_i}{{\bm{y}}_i}} \! \mid \! \sum {{a_i}}  = 1,0 \le {a_i} \le \tau } \right\}
\end{array}
\label{ACLC}
\end{eqnarray}
If ${\tau } =  1$, $H(\bm{Y})$ is a convex hull \cite{rockafellar1997convex}.
If ${\tau }<1$, $H(\bm{Y})$ is a reduced convex hull \cite{burges2000geometric}.
For the convenience of expression, in the following the development we call both the cases convex hull.

\begin{figure}[h]
\centering
\includegraphics[height=3cm]{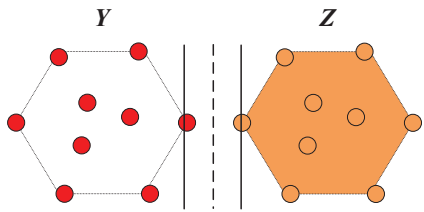}
\caption{Convex hull based set to set distance.}
\label{svm}
\end{figure}

By modeling a set as a convex hull, the distance between set $\bm{Y} = \{ {\bm{y}_1},...,{\bm{y}_i},...,{\bm{y}_{n_1}}\}$ and set $\bm{Z} = \{ {\bm{z}_1},...,{\bm{z}_j},...,{\bm{z}_{n_2}}\}$ can be defined as follows:
\begin{eqnarray}
 \begin{array}{l}
 {\min _{{\bm{a}},{\bm{b}}}}\left\| {\sum {{a_i}{{\bm{y}}_i} - \sum {{b_j}{{\bm{z}}_j}} } } \right\|_2^2 \\
 s.t.\sum {{a_i}}  = 1,0 \le {a_i} \le \tau  \\
 \quad  \; \; \sum {{b_j}}  = 1,0 \le {b_j} \le \tau  \\
 \end{array}
 \label{settoset}
\end{eqnarray}
 When the two sets have no intersection, the set to set distance in Eq. (\ref{settoset}) becomes the distance between the nearest points in the two convex hulls (CHISD \cite{cevikalp2010face}), as illustrated in Fig. \ref{svm}.
 It is not difficult to see that such a distance is equivalent to the distance computed by SVM \cite{burges2000geometric}.
 If the discriminative function of SVM is $f=\bm{wx}+b$, then $\bm{w}= {\sum {{a_i}{{\bm{y}}_i} - \sum {{b_j}{{\bm{z}}_j}} } }$ and the margin is $2/\|\bm{w}\|$.
If we consider each image set as one class, then maximizing margin between the two classes is equivalent to finding the set to set distance \cite{cevikalp2010large}.
However, such a distance relies highly on the location of each individual sample and can be sensitive to outliers \cite{wang2012covariance}.

\subsection{Collaborative representation based set to sets distance and classification}
\label{section22}
In image set based face recognition (ISFR), we have a query set $\bm{Y}$ but multiple gallery sets ${\bm{X}}_k, k=1,2,...,K$. One fact in face recognition is that the face images from different people still have much similarity. If we compute the distance between $\bm{Y}$ and each ${\bm{X}}_k$ by using methods such as hull based set to set distance (refer to \ref{section21}), the correlation between different gallery sets will not be utilized. As we discussed in the Introduction section, inspired by the SRC \cite{wright2009robust} and CRC \cite{zhang2011sparse} methods in image based face recognition, here we propose a novel ISFR method, namely image set based collaborative representation and classification (ISCRC).

The key component of ISCRC is the collaborative representation based set to sets distance (CRSSD) defined as follows. Let ${\bm{X}} = [{{\bm{X}}_1},...,{{\bm{X}}_k},...,{{\bm{X}}_K}]$ be the concatenation of all gallery sets. We model each of $\bm{Y}$ and  $\bm{X}$ as a hull, i.e., ${\bm{Y}}{\bm{a}}$ and ${\bm{X}}{\bm{b}}$, where $\bm{a}$ and  $\bm{b}$ are coefficient vectors, and then we define the CRSSD between set $\bm{Y}$ and sets $\bm{X}$ as:
\begin{eqnarray}
\begin{array}{l}
 {\min _{\bm{a,b}}}\left\| {{\bm{Ya}} - {\bm{Xb}}} \right\|_{}^2 \; s.t.\sum\nolimits {{a_i}}  = 1 \\
 \end{array}
 \label{ssd}
\end{eqnarray}
where $a_i$ is the $i^{th}$ coefficeint in $\bm{a}$ and we let $\sum {{a_i}}  = 1$ to avoid the trivial solution $\bm{a}=\bm{b}= \bm{0}$.  In Eq. (\ref{ssd}), the hull ${\bm{Y}}{\bm{a}}$ of the query set ${\bm{Y}}$ is collaboratively represented over the gallery sets; however, the coefficients in $\bm{a}$ will make the samples in $\bm{Y}$ be treated differently in the representation and the subsequent classification process.

Suppose that the coefficient vectors $\bm{\hat a}$ and $\bm{\hat b}$ are obtained by solving Eq. (\ref{ssd}), then we can write $\bm{\hat b}$ as ${\bm{\hat b}} = [{{\bm{\hat b}}_1};...;{{\bm{\hat b}}_k};...;{{\bm{\hat b}}_K}]$, where ${\bm{\hat b}}_k$ is is the sub-vector of coefficients associated with gallery set ${\bm{X}}_k$.
Similar to the classification in SRC and CRC, we use the representation residual of hull $\bm{Y}\bm{\hat a}$ by each set ${\bm{X}}_k$ to determine the class label of $\bm{Y}$. The classifier in the proposed ISCRC is:
\begin{eqnarray}
Identity(\bm{Y})=argmin_k \left\{ {r_k} \right\}
\label{ICSRC}
\end{eqnarray}
where ${r_k} = \left\| {\bm{Y {\hat a}} - {{\bm{X}}_k}{{\bm{\hat b}}_k}} \right\|_2^2$.

Clearly, the solutions to $\bm{a}$ and $\bm{b}$ in Eq. (\ref{ssd}) determine the CRSSD and hence the result of ISCRC. In order to get stable solutions, we could impose reasonable regularizations on $\bm{a}$ and $\bm{b}$. In the following sections \ref{section23} and \ref{section24}, we discuss the convex hull based CRSSD and regularized hull based CRSSD, respectively.

\subsection{Convex hull based CRSSD}
\label{section23}
One important instantiation of CRSSD is the convex hull based CRSSD. In this case, both the hulls $\bm{Ya}$ and $\bm{Xb}$ are required to be convex hulls, and then the distance in Eq. (\ref{ssd}) becomes
\begin{eqnarray}
\begin{array}{l}
 {\min _{\bm{a,b}}}\left\| {{\bm{Ya}} - {\bm{Xb}}} \right\|_{}^2 \\
 s.t.\sum\limits{{a_i}}  = 1,\sum\limits{{b_j}}  = 1, \\
 \quad \; \; 0 \le {a_i} \le {\tau },i = 1,...,n_a,\\
 \quad \; \; 0 \le {b_j} \le {\tau },j = 1,...,n_b \\
 \end{array}
 \label{ALC}
\end{eqnarray}
where $a_i$ and $b_j$ are the $i^{th}$ and $j^{th}$ coefficients in $\bm{a}$ and $\bm{b}$, respectively, $n_a$ and $n_b$ are the number of samples in set $\bm{Y}$ and sets $\bm{X}$, respectively, and $\tau \le 1$.
\begin{figure}[h]
\centering
\includegraphics[height=5cm]{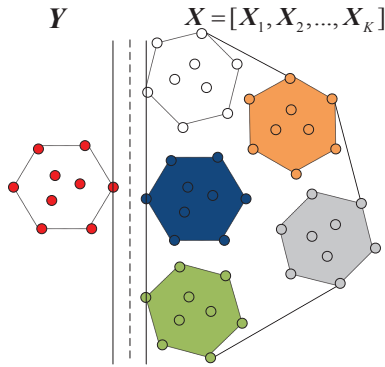}
\caption{Convex hull based CRSSD.}
\label{CHC}
\end{figure}

A geometric illustration of convex hull based CRSSD is shown in Fig. \ref{CHC}. Different from the CHISD method in \cite{cevikalp2010face}, which models each gallery set as a convex hull, here we model all the gallery sets as one big convex hull. Similar to the closest points searching in SVM, convex hull based CRSSD aims to find the closest points in the query set $\bm{Y}$ and the whole gallery set $\bm{X}$ in a large margin manner. With convex hull based CRSSD, the corresponding ISCRC method can be viewed as a large margin based classifier in some sense. Nonetheless, the classification rules in SVM and ISCRC are very different.

\subsection{$l_p$-norm regularized hull based CRSSD}
\label{section24}

The convex hull modeling of a set can be affected much by outlier samples in the set \cite{wang2012covariance}. To make CRSSD more stable, the  $l_p$-norm regularized hull can be used to model $\bm{Y}$ and $\bm{X}$. For the query set $\bm{Y}$, we should keep the constraint $\sum {{a_i}}  = 1$ to avoid the trivial solution, and the $l_p$-norm regularized hull of $\bm{Y}$ is defined as
\begin{eqnarray}
 \begin{array}{l}
H(\bm{Y}) = \{ \sum {{a_i}{\bm{y}_i}} \! \mid \! {\left\| \bm{a} \right\|_{l_p}} < \delta \} \; s.t. \sum {{a_i}}  = 1\\
 \end{array}
\label{RHY}
\end{eqnarray}
For the gallery set $\bm{X}$, its regularized hull is defined as:
\begin{eqnarray}
 \begin{array}{l}
H(\bm{X}) = \{ \sum {{b_i}{\bm{x}_i}} \!\mid \! {\left\| \bm{b} \right\|_{l_p}} < \delta \}\\
 \end{array}
\label{RHX}
\end{eqnarray}
Finally, the regularized hull based CRSSD between $\bm{Y}$ and $\bm{X}$ is defined as:
 \begin{eqnarray}
 \begin{array}{l}
 {\min _{{\bm{a,b}}}}\left\| {{\bm{Ya}} - {\bm{Xb}}} \right\|_2^2 \\
 s.t. {\left\| \bm{a} \right\|_{l_p}} < {\delta _1},{\left\| \bm{b} \right\|_{l_p}} < {\delta _2},\sum {{a_i}}  = 1 \\
 \end{array}
 \label{SLC1}
 \end{eqnarray}

\section{Regularized hull based ISCRC}
\label{section3}
In Section \ref{section2}, we introduced CRSSD, and presented two important instantiations of it, i.e., convex hull based CRSSD and regularized hull based CRSSD. With either one of them, the ISCRC (refer to Eq. (\ref{ICSRC})) can be implemented to perform ISFR. In this section, we discuss the minimization of regularized hull based CRSSD model, and the corresponding classification scheme is called regularized hull based ISCRC, denoted by RH-ISCRC. The minimization of convex hull based CRSSD and the corresponding classification scheme will be discussed in Section \ref{section4}.

\subsection{Main model}
We can re-write the regularized hull based CRSSD model in Eq. (\ref{SLC1}) as its Lagrangian formulation:
 \begin{eqnarray}
\begin{array}{l}
 \min_{\bm{a},\bm{b}} \left\| {\bm{Ya} - \bm{Xb}} \right\|_2^2 + {\lambda _1}{\left\| \bm{a} \right\|_{l_p}} + {\lambda _2}{\left\| \bm{b} \right\|_{l_p}} \\
 s.t.\sum\nolimits {{a_i}}  = 1 \\
 \end{array}
 \label{slc2}
\end{eqnarray}
where ${\lambda _1}$ and ${\lambda _2}$ are positive constants to balance the representation residual and the regularizer.

In ISFR, each gallery set ${\bm{X}}_k$ often has tens to hundreds of sample images so that the whole set $\bm{X}$ can be very big, making the computational cost to solve Eq. (\ref{slc2}) very high. Considering the fact that the images in each set ${\bm{X}}_k$ have high redundancy, we can compress ${\bm{X}}_k$ into a much more compact set, denoted by ${\bm{D}}_k$, via dictionary learning methods such as KSVD \cite{rubinstein2008efficient} and metaface learning \cite{yang2010metaface}. Let ${\bm{D}} = [{{\bm{D}}_1},...,{{\bm{D}}_k},...,{{\bm{D}}_K}]$. We can then replace $\bm{X}$ by $\bm{D}$ in Eq. (\ref{slc2}) to compute the regularized hull based CRSSD:
 \begin{eqnarray}
\begin{array}{l}
({\bm{\hat a}},{\bm{\hat \beta }}) = \arg {\min _{{\bm{a}},{\bm{\beta }}}}\left\{ \begin{array}{l}
\left\| {{\bm{Ya}} - {\bm{D {\beta} }}} \right\|_2^2 + \\
{\lambda _1}{\left\| {\bm{a}} \right\|_{{l_p}}} + {\lambda _2}{\left\| {\bm{\beta }} \right\|_{{l_p}}}
\end{array} \right\}\\
s.t.\sum\nolimits {{a_i}}  = 1
\end{array}
 \label{slc3}
\end{eqnarray}
where ${\bm{\beta}} = [{{\bm{\beta}}_1};...;{{\bm{\beta}}_k};...;{{\bm{\beta}}_K}]$ and ${\bm{\beta}}_k$ is the sub-vector of coefficients associated with ${\bm{D}}_k$.
Based on our experimental results, compressing ${\bm{X}}_k$ into ${\bm{D}}_k$ significantly improve the speed with almost the same ISFR rate.

Either $l_1$-norm or $l_2$-norm can be used to regularize $\bm{a}$ and $\bm{\beta}$, while $l_1$-regularization will lead to sparser solutions but with more computational cost. Like in  $l_1$-SVM \cite{zhu20041} and SRC \cite{wright2009robust}, sparsity can enhance the classification rate if the features are not informative enough.  Note that if the query set $\bm{Y}$ has only one sample, then $\bm{a}=[1]$ and the proposed model in Eq. (\ref{slc3}) will be reduced to the SRC (for $l_1$-regularization) or CRC (for $l_2$-regularization) scheme.
Next, we present the optimization of $l_2$-norm and $l_1$-norm regularized hull based ISCRC in Section \ref{l2rchiscrc} and Section \ref{l1rchiscrc}, respectively.

\subsection{$l_2$-norm regularized hull based ISCRC}
\label{l2rchiscrc}
 When $l_2$-norm is used to regularize $\bm{a}$ and $\bm{\beta}$, the problem in Eq. (\ref{slc3}) has a closed-form solution. The Lagrangian function of Eq. (\ref{slc3}) becomes
\begin{eqnarray}
\begin{array}{l}
L({\bm{a}},{\bm{\beta }},{\lambda _3}) = \left\| {{\bm{Ya}} - {\bm{D\beta }}} \right\|_2^2 + {\lambda _1}\left\| {\bm{a}} \right\|_2^2 + {\lambda _2}\left\| {\bm{\beta }} \right\|_2^2 + {\lambda _3}({\bm{ea}} -1)\\
 = \left\| {\left[ {{\bm{Y}}{\rm{ }} \;-{\bm{D}}} \right]\left[ \begin{array}{l}
{\bm{a}}\\
{\bm{\beta }}
\end{array} \right]} \right\|_2^2 + \left[ {{{\bm{a}}^T}{\rm{ }}\; \;{{\bm{\beta }}^T}} \right]\left[ \begin{array}{l}
{\lambda _1}{\bm{I}}{\rm{    }}\; \; \bm{0}\\
{\rm{ }}\bm{0} \; \; {\rm{    }}{\lambda _2}{\bm{I}}
\end{array} \right]\left[ \begin{array}{l}
{\bm{a}}\\
{\bm{\beta }}
\end{array} \right]
 + {\lambda _3}(\left[ {{\bm{e}}\; \;{\rm{ }}{\bm{0}}} \right]\left[ \begin{array}{l}
{\bm{a}}\\
{\bm{\beta }}
\end{array} \right] - 1)
\end{array}
\label{rhl2}
\end{eqnarray}
where $\bm{e}$ is a row vector whose elements are 1.

Let ${\bm{z}} = \left[ \begin{array}{l}
{\bm{a}}\\
{\bm{\beta }}
\end{array} \right]$,
$\bm{A}={\left[ {{\bm{Y}}{\rm{ }}\;\; - {\bm{D}}} \right]}$,
$\bm{B}=\left[ \begin{array}{l}
{\lambda _1}{\bm{I}}{\rm{    }}\;\;\bm{0}\\
{\rm{ }}\bm{0}\;\;{\rm{    }}{\lambda _2}{\bm{I}}
\end{array} \right]$ and ${\bm{d}} = {\left[ {{\bm{e}}{\rm{ }}\;\;{\bm{0}}} \right]^T}$.
Then Eq. (\ref{rhl2}) becomes:
 \begin{eqnarray}
L({\bm{z}},{\lambda _3}) = {{\bm{z}}^T}{{\bm{A}}^T}{\bm{Az}} + {{\bm{z}}^T}{\bm{Bz}} + {\lambda _3}({{\bm{d}}^T}{\bm{z}}-1)
 \label{rhl2z}
\end{eqnarray}
There are
\begin{eqnarray}
\frac{{\partial L}}{{\partial {\lambda _3}}} = {{\bm{d}}^T}{\bm{z}}-1 = 0
\label{rhl2z1}
\end{eqnarray}
\begin{eqnarray}
\frac{{\partial L}}{{\partial z}} = {{\bm{A}}^T}{\bm{Az}} + {\bm{Bz}} + {\lambda _3}{\bm{d}} = 0
 \label{rhl2z2}
\end{eqnarray}
According to Eq. (\ref{rhl2z1}) and Eq. (\ref{rhl2z2}), we get the closed form solution to Eq. (\ref{rhl2}):
\begin{eqnarray}
{\bm{\hat z}} = \left[ \begin{array}{l}
{{\bm{\hat a}}}\\
{{\bm{\hat \beta }}}
\end{array} \right]={{\bm{z}}_0}/{{\bm{d}}^T}{{\bm{z}}_0}
 \label{rhl2solution}
\end{eqnarray}
where ${{\bm{z}}_0} = {({{\bm{A}}^T}{\bm{A}} + {\bm{B}})^{ - 1}}\bm{d}$.

After $\bm{\hat a}$ and $\bm{\hat {\beta}}$ are got, the distance between query set $\bm{Y}$ and a gallery set ${\bm{X}}_k$ is calculated as ${r_k} = \left\| {{\bm{Y {\hat a}}} - {{\bm{D}}_k}{{\bm{\hat {\beta}}}_k}} \right\|_2^2$, and then the class label of $\bm{Y}$ is determined by Eq. (\ref{ICSRC}).
For RH-ISCRC-$l_2$, the main time consumption is to solve the inverse of matrix ${({{\bm{A}}^T}{\bm{A}} + {\bm{B}})}$.
Hence, the time complexity of RH-ISCRC-$l_2$ is $O({\left( {{n_a }}+{n_{\beta}} \right)^3})$, where $n_a$ is the number of sample images in $\bm{Y}$ and $n_{\beta}$ is the number of atoms in $\bm{D}$.

\subsection{$l_1$-norm regularized hull based ISCRC}
\label{l1rchiscrc}
When $l_1$-norm regularization is used, we use the alternating minimization method, which is very efficient to solve multiple variable optimization problems \cite{gunawardana2005convergence}.
For Eq. (\ref{slc3}), we have the following augmented Lagrangian function:
 \begin{eqnarray}
\begin{array}{l}
 L(\bm{a},\bm{\beta},\lambda ) = \left\| {\bm{Ya} - \bm{D{\beta}}} \right\|_2^2 + {\lambda _1}{\left\| \bm{a} \right\|_{1}} + {\lambda _2}{\left\| \bm{\beta} \right\|_{1}}  \\
  + < \lambda ,\bm{ea}- 1>  + \frac{\gamma }{2}\left\| {\bm{ea} - 1} \right\|_2^2 \\
 \end{array}
 \label{admover}
\end{eqnarray}
where $\lambda$ is the Lagrange multiplier, $\langle \cdot,\cdot \rangle$ is the inner product, and $\gamma> 0$ is the penalty parameter.

Then $\bm{a}$ and $\bm{\beta}$ are optimized alternatively with the other one fixed.
More specifically, the iterations of minimizing $\bm{a}$ go as follows:
\begin{eqnarray}
\begin{array}{l}
 {{\bm{a}^{(t + 1)}}} = \arg {\min _{\bm{a}}}L(\bm{a},{\bm{\beta}^{(t)}},{\lambda^{(t)}}) \\
  \quad \quad \; \; \;= \arg {\min _{\bm{a}}}f(\bm{a}) + \frac{\gamma }{2}\left\| {\bm{ea} - 1 + {\lambda ^{(t)}}/\gamma } \right\|_2^2 \\
  \quad \quad \; \; \;= \arg {\min _{\bm{a}}}\left\| {{{\tilde {\bm Y}} {\bm{a}}} - \bm{x}} \right\|_2^2 + {\lambda _1}{\left\| \bm{a} \right\|_{1}} \\
 \end{array}
 \label{ADMa}
 \end{eqnarray}
 where $f(\bm{a}) = \left\| {\bm{Ya} - \bm{D}{{\bm{\beta}}^{(t)}}} \right\|_2^2 + {\lambda _1}{\left\| \bm{a} \right\|_{l_p}}$,
 $\bm{\tilde Y} = \left[ {{\bm{Y}};{{(\gamma /2)}^{1/2}}\bm{e}} \right]$,
 $\bm{x} = [\bm{D}{{\bm{\beta}}^{(t)}};{(\gamma /2)^{1/2}}(1 - {\lambda ^{(t)}}/\gamma )]$.

 The problem in Eq. (\ref{ADMa}) can be easily solved by some representative $l_1$-minimization approaches \cite{TIPSR} such as LARS \cite{efron2004least}.

After $\bm{a}^{(t + 1)}$ is updated, $\bm{\beta}^{(t + 1)}$ can be obtained by solving another $l_1$-regularized optimization problem:
 \begin{eqnarray}
 \begin{array}{l}
 {{\bm{\beta}}^{(t + 1)}} = \arg {\min _{\bm{\beta}}}L({{\bm{a}}^{(t+1)}},\bm{\beta},{\lambda^t}) \\
 \quad \quad \; \; \; \;  = \arg {\min _{\bm{\beta}}}\left\| {\bm{Y}{\bm{a}^{(t+1)}} - \bm{D{\beta}}} \right\|_2^2 + {\lambda _2}{\left\| \bm{\beta} \right\|_{1}} \\
   \end{array}
   \label{ADMb}
 \end{eqnarray}
Once ${\bm{a}}^{(t + 1)}$ and ${\bm{\beta}}^{(t + 1)}$ are got, $\lambda$ is updated as follows:
 \begin{eqnarray}
 {\lambda ^{(t + 1)}} = {\lambda ^{(t)}} + \gamma \left( {\bm{e}{{\bm{a}}^{(t + 1)}} - 1} \right)
   \label{ADMlamda}
  \end{eqnarray}

\begin{figure}[h]
\centering
\includegraphics[height=4cm]{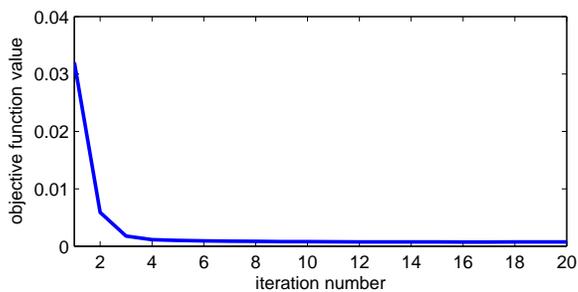}
\caption{Convergence of RH-ISCRC-$l_1$.}
\label{converrch}
\end{figure}

The algorithm of RH-ISCRC-$l_1$ for ISFR is summarized in Table \ref{AlgorithmSAH} and it converges.
The problem in Eq. (\ref{admover}) is convex, and the subproblems in Eq. (\ref{ADMa}) and Eq. (\ref{ADMb}) are convex and can be solved using the LARS algorithm.
It had been shown in \cite{niesen2009adaptive}, for the general convex problem, the alternating minimization approach would converge to the correct solution.
One curve of the objective function value of RH-ISCRC-$l_1$ versus the iteration number is shown in Fig. \ref{converrch}, where the Honda/USCD\footnote{http://vision.ucsd.edu/~leekc/HondaUCSDVideoDatabase/HondaUCSD.html} database \cite{lee2003video} is used.
The query set $\bm{Y}$ and each gallery set ${\bm{X}}_k$ has 200 frames, and we compress each set ${\bm{X}}_k$  into a dictionary ${\bm{D}}_k$ with 20 atoms by using the metaface learning method \cite{yang2010metaface}.
Since there are 20 gallery sets, the set ${\bm{D}} = [{{\bm{D}}_1},...,{{\bm{D}}_k},...,{{\bm{D}}_{20}}]$ has 20 $\times$ 20=400 atoms.
From the figure we can see that RH-ISCRC-$l_1$ converges after about five iterations.

\begin{table}[h]
\begin{center}
\caption {Algorithm of RH-ISCRC for ISFR}
\label{AlgorithmSAH}
\tabcolsep=0.1in
\begin{tabular}{l}
\hline
\textbf{Input}: query set $\bm{Y}$; gallery sets ${\bm{X}} = [{{\bm{X}}_1},...,{{\bm{X}}_k},...,{{\bm{X}}_K}]$, $\lambda_1$ and $\lambda_2$.\\
\textbf{Output}: the label of query set $\bm{Y}$.\\
Initialize ${\beta}^{(0)}$, $\lambda^{(0)}$ and $0\leftarrow t$.\\
Compress ${\bm{X}}_k$ to ${\bm{D}}_k$, $k=1,2,...,K$ using metaface learning \cite{yang2010metaface}.\\
While $t<max\_num$ do\\
\quad Step 1: Update ${\bm{a}}$ by Eq. (\ref{ADMa});\\
\quad Step 2: Update ${\bm{\beta}}$ by Eq. (\ref{ADMb});\\
\quad Step 3: Update $\lambda$ by Eq. (\ref{ADMlamda});\\
\quad Step 4: $t \leftarrow t + 1$.\\
End while\\
Compute ${r_k} = \left\| {{\bm{Y {\hat a}}} - {{\bm{D}}_k}{{\bm{\hat {\beta}}}_k}} \right\|_2^2$, $k=1,2,...K.$\\
Identity(${\bm{Y}}$)=$\arg {\min _k}\{ {r_k}\}$.\\
\hline
\end{tabular}
\end{center}
\end{table}
 Since the complexity of sparse coding is $O({m^2}{{n}^\varepsilon })$, where $m$ is the feature dimension, $n$ is the atom number and $\varepsilon\geq1.2$ \cite{kim2007interior}, we can get that the time complexity of RH-ISCRC-$l_1$ is $O(l{m^2}{({n_a}^\varepsilon +{n_{\beta}}^\varepsilon )})$, where $n_a$ is the number of samples in $\bm{Y}$, $n_{\beta}$ is the number of atoms in $\bm{D}$ and $l$ is the iteration number.

\subsection{Examples and discussions}
\begin{figure}[h]
\centering
\includegraphics[height=4cm]{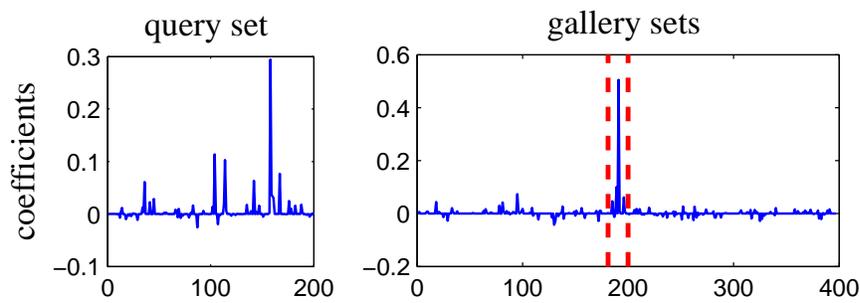}
\caption{The coefficient vectors $\bm{\hat a}$ (of $\bm{Y}$) and $\bm{\hat{\beta}}$ (of $\bm{D}$) by $l_1$-regularized hull based CRSSD.}
\label{XaDb1}
\end{figure}

\begin{figure}[h]
\centering
\includegraphics[height=4cm]{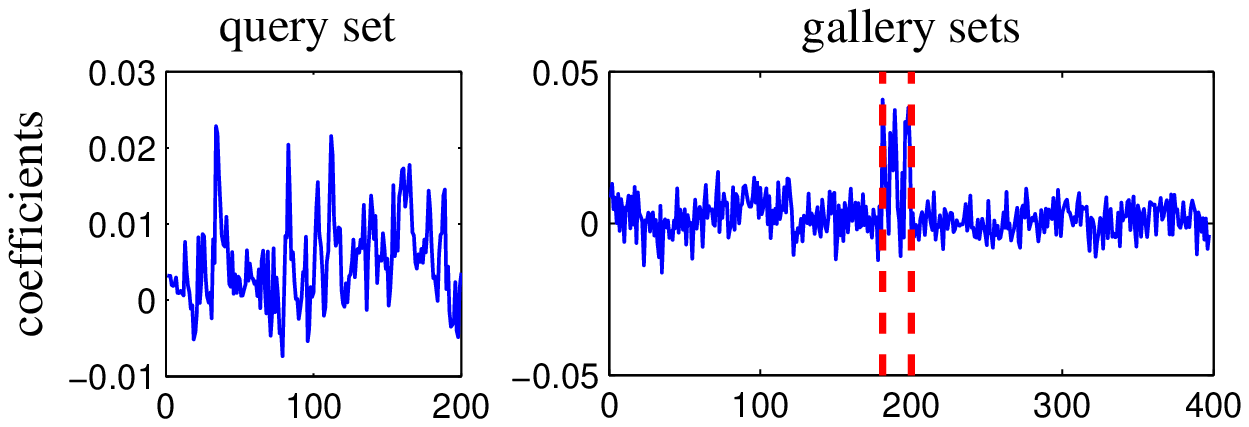}
\caption{The coefficient vectors $\bm{\hat a}$ (of $\bm{Y}$) and $\bm{\hat{\beta}}$ (of $\bm{D}$) by $l_2$-regularized hull based CRSSD.}
\label{XaDb2}
\end{figure}

Let's use an example to better illustrate the classification process of RH-ISCRC.
We use the Honda/USCD database \cite{lee2003video}.
The experiment setting is the same as Fig. \ref{converrch}.
By Eq. (\ref{slc3}), the computed coefficients in $\bm{a}$ and $\bm{\beta}$ are plotted in Fig. \ref{XaDb1} (by $l_1$-regularization) and Fig. \ref{XaDb2} (by $l_2$-regularization), respectively. The highlighted coefficients in the figures are associated with set  ${\bm{X}}_{10}$, which has the same class label as $\bm{Y}$. Clearly, these coefficients are much more significant than the coefficients associated with the other classes. Meanwhile, from Fig. \ref{XaDb1} and Fig. \ref{XaDb2} we can see that $l_1$-regularized hull based CRSSD leads to sparser $\bm{a}$ and $\bm{\beta}$, implying that only few samples are dominantly involved in representation and classification.

\begin{figure}[h]
\centering
\includegraphics[width=10cm]{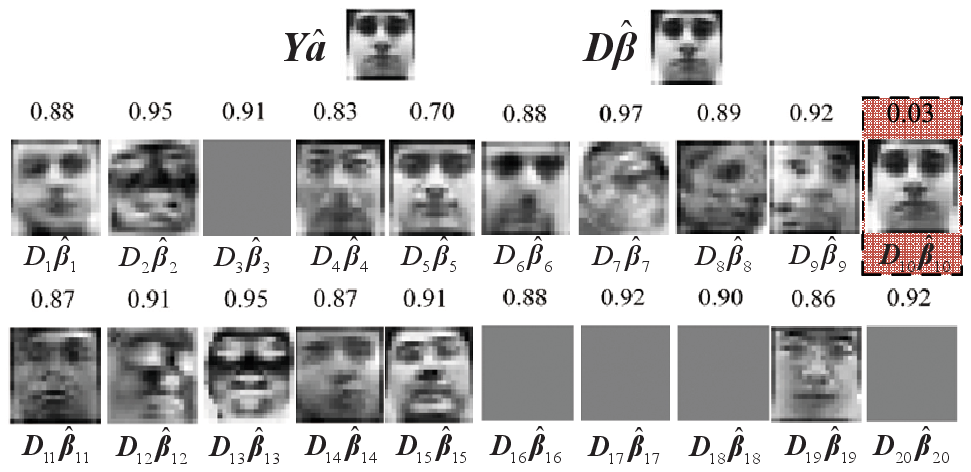}
\caption{Reconstructed faces $\bm{Y {\hat{a}}}$, $\bm{D{\hat{\beta}}}$, ${\bm{D}}_k {\bm{\hat{\beta}}}_k$ (we normalized each ${\bm{D}}_k {\bm{\hat{\beta}}}_k$ for better visualization).
The number over each ${\bm{D}}_k {\bm{\hat{\beta}}}_k$ is the residual ${r_k} = \left\| {{\bm{Y {\hat{a}}}} - {{\bm{D}}_k}{{\bm{\hat{\beta}}}_k}} \right\|_2^2$.}
\label{XD}
\end{figure}

In Fig. \ref{XD}, we show the reconstructed faces by $\bm{Y {\hat a}}$ with $l_1$-regularized hull based CRSSD. The distances between $\bm{Y{\hat a}}$  and each ${\bm{D}}_k{{\bm{\hat{\beta}}}_k}$, i.e., $r_k$, are also given. We see that $r_{10}$ is 0.03, which is the minimal one among all the gallery sets, meaning that ISCRC will make the correct recognition.
Here the relationships between ISCRC and manifold based methods can be revealed.
MMD assumes that an image set can be modeled as a set of local subspaces so that the image set distance is defined as the weighted average distance between any two local subspaces \cite{wang2008manifold}.
The distance between two local subspaces is related to the cluster exemplar and principle angel.
Correspondingly, ISCRC seeks for a local subspace ($\bm{Y{\hat a}}$) in the query image set and a local subspace (${\bm{D}}{{\bm{\hat{\beta}}}}$) in all the gallery sets, as shown in Fig. \ref{XaDb1} .
In classification, the distance between the query set and the template set of the $k^{th}$ class
is the distance between the local subspace ($\bm{Y{\hat a}}$) and the local subspace ${\bm{D}}_k{{\bm{\hat{\beta}}}_k}$.

\section{Kernelized convex hull based ISCRC}
\label{section4}
We then focus on how to compute the convex hull based CRSSD in Eq. (\ref{ALC}) and use it for ISCRC.
Since there can be many sample images in gallery sets, $\bm{X}$ can be a fat matrix (note that usually we use a low dimensional feature vector to represent each face image). Even we compress $\bm{X}$ into a more compact set $\bm{D}$, the system can still be under-determined. In Section 3 we imposed the $l_p$-norm regularization on $\bm{a}$ and $\bm{b}$ to make the solution stable. When the convex hull is used, however, the constraint may not be strong enough to get a stable solution of Eq. (\ref{ALC}). In addition, if the underlying relationship between the query set and gallery sets is highly nonlinear, it is difficult to approximate the hull of query set as a linear combination of gallery sets.

One simple solution to solving both the above two problems is the kernel trick; that is, we can map the data into a higher dimensional space where the subjects can be approximately linearly separable. The mapped gallery data matrix in the high-dimensional space will be generally over-determined. In such a case, the convex hull constraint will be strong enough for a stable solution. The kernelized convex hull based CRSSD model is:
 \begin{eqnarray}
\begin{array}{l}
 {\min _{\bm{a,{\beta}}}}\left\| {\phi ({\bm{Y}}){\bm{a}} - \left[ {\phi ({{\bm{D}}_1}),\phi ({{\bm{D}}_2}),...,\phi ({{\bm{D}}_K})} \right]{\bm{\beta}}} \right\|_{}^2 \\
 s.t.\sum\limits{{a_i}}  = 1,\sum\limits{{{\beta}_j}}  = 1, \\
 \quad \; \; 0 \le {a_i} \le {\tau},i = 1,...,n_a,\\
 \quad \; \; 0 \le {{\beta}_j} \le {\tau},j = 1,...,n_{\beta}. \\
 \end{array}
 \label{ALCK}
 \end{eqnarray}

The above minimization can be easily solved by the standard quadratic optimization (QP \cite{coleman1996reflective}) method.
The solution exhibits global and quadratic convergence, as proved in \cite{coleman1996reflective}.
Different kernel functions can be used, e.g., linear kernel and Gaussian kernel.
We call the corresponding method kernelized convex hull based ISCRC, denoted by KCH-ISCRC. The classification rule is the same as RH-ISCRC with ${r_k} = \left\| {\phi ({\bm{Y}}){\bm{\hat a}} - \phi ({{\bm{D}}_k}){{\bm{\hat{\beta}}}_k}} \right\|{_2^2}$. As convex hull based CRSSD is to solve a convex QP problem, the time complexity of KCH-ISCRC is $O((n_{\beta}+n_a)^3)$, which is similar to SVM.
The algorithm of KCH-ISCRC is given in Table \ref{AlgorithmKCH}.
To reduce the computational cost, the kernel matrix $k(\bm{D},\bm{D})$ can be computed and stored.
When a query set $\bm{Y}$ comes, we only need to calculate $k(\bm{Y},\bm{Y})$ and $k(\bm{Y},\bm{D})$.
\begin{table}[h]
\begin{center}
\caption {Algorithm of KCH-ISCRC for ISFR}
\label{AlgorithmKCH}
\tabcolsep=0.1in
\begin{tabular}{l}
\hline
\textbf{Input}: query set $\bm{Y}$; gallery sets ${\bm{X}} = [{{\bm{X}}_1},..., {{\bm{X}}_k},...,{{\bm{X}}_K}]$, $\tau$.\\
\textbf{Output}: the label of query set $\bm{Y}$.\\
Compress ${\bm{X}}_k$ to ${\bm{D}}_k$, $k=1,2,...,K$ by meaface learning \cite{yang2013face};\\
Solve the QP problem in Eq. (\ref{ALCK});\\
Compute ${r_k} = \left\| {\phi ({\bm{Y}}){\bm{\hat a}} - \phi ({{\bm{D}}_k}){{\bm{\hat{\beta}}}_k}} \right\|{_2^2}$, $k=1,2,...K;$\\
Identity(${\bm{Y}}$)=$\arg {\min _k}\{ {r_k}\}$.\\
\hline
\end{tabular}
\end{center}
\end{table}

\begin{figure}[h]
\centering
\includegraphics[height=4cm]{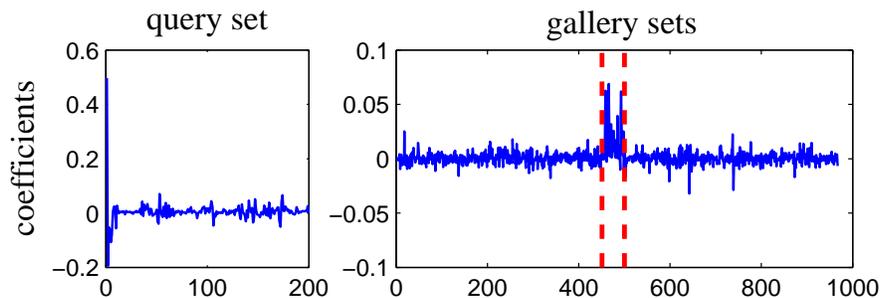}
\caption{The coefficient vectors $\bm{\hat a}$ (of $\bm{Y}$) and $\bm{\hat{\beta}}$ (of $\bm{D}$) by kernelized convex hull based CRSSD.}
\label{XaDbC}
\end{figure}

Like in Fig. \ref{XaDb1} and Fig. \ref{XaDb2}, in Fig. \ref{XaDbC} we show the coefficient vectors $\bm{\hat a}$ and $\bm{\hat{\beta}}$ solved by Eq. (\ref{ALCK}). The Gaussian kernel is used and the experimental setting is the same as that in Figs. \ref{XaDb1} and \ref{XaDb2} (the only difference is that each compressed gallery set ${\bm{D}}_k$ has 50 atoms). We can see that the coefficients associated with gallery set ${\bm{D}}_{10}$ are larger than the other gallery sets, resulting in a smaller representation residual and hence the correct recognition.

\section{Experimental analysis}
\label{section5}
We used the Honda/UCSD \cite{lee2003video}, CMU Mobo \cite{gross2001cmu}, and Youtube Celebrities \cite{kim2008face} datasets to
test the performance of the proposed method.
The comparison methods fall into four categories:
\begin{itemize}
 \item[C1.] Subspace and manifold based methods: Mutual Subspace Method (MSM) \cite{yamaguchi1998face}, Discriminant Canonical Correlations (DCC\footnote{http://www.iis.ee.ic.ac.uk/~tkkim/code.htm}) \cite{kim2007discriminative}, Manifold-Manifold Distance (MMD\footnote{http://www.jdl.ac.cn/user/rpwang/research.htm}) \cite{wang2008manifold}, and Manifold Discriminant Analysis (MDA\footnote{http://www.jdl.ac.cn/user/rpwang/research.htm}) \cite{wang2009manifold}.
 \item[C2.] Affine/convex hull based methods: Affine Hull based Image Set Distance (AHISD\footnote{http://www2.ogu.edu.tr/~mlcv/softwareimageset.html}) \cite{cevikalp2010face}, Convex Hull based Image Set Distance (CHISD\footnote{http://www2.ogu.edu.tr/~mlcv/softwareimageset.html}) \cite{cevikalp2010face}, Sparse Approximated Nearest Points (SANP\footnote{https://sites.google.com/site/yiqunhu/cresearch/sanp}) \cite{hu2011sparse}, and Regularized Nearest Points (RNP) \cite{yang2013face}.
 \item[C3.] Representation based methods: Sparse Representation based Classifier (SRC) \cite{wright2009robust}, Collaborative Representation based Classifier (CRC) \cite{zhang2011sparse}. We tested to use the average and minimal representation residual of query set for classification and found that average residual works better. Hence in this paper, the average residual is used in SRC/CRC for classification.
 \item[C4.] Kernel methods: KSRC (Kernel SRC) \cite{gao2010kernel}, KCRC (Kernel CRC) \cite{yang2013}, AHISD \cite{cevikalp2010face}, and CHISD \cite{cevikalp2010face}. For KSRC and KCRC, the average residual is used for classification.
\end{itemize}

For the proposed methods, RH-ISCRC is compared with those non-kernel methods and KCH-ISCRC is compared with those kernel methods.
\subsection{Parameter setting}
For competing methods, the important parameters were empirically tuned according to the recommendations in the original literature for fair comparison.
For DCC \cite{kim2007discriminative}, if there is only one set per class, then the training set is divided into two sets since at least two sets per class are needed in DCC.
For MMD, the number of local models is set following the work in \cite{wang2008manifold}.
For MDA, there are three parameters, i.e., the number of local models, the number of between-class NN local
models and the subspace dimension.
The three parameters are configured according to the work in \cite{wang2009manifold}.
For SANP, we adopted the same parameters as \cite{hu2011sparse}.
For SRC, CRC, KSRC and KCRC, $\lambda$ that balances the residual and regularization is tuned from $[0.01, 0.001, 0.0001]$.
For AHISD and CHISD, $C$ is set as 100.
For all kernel methods, Gaussian kernel ($k(x,y) = \exp ( - \left\| {x - y} \right\|_2^2/2{\delta ^2})$) is used, and $\delta$ is set as 5.
The experiments of 50 frames, 100 frames and 200 frames per set are conducted on the three databases.
If the number of samples in the set is less than the given number, then all the samples in the set are used.

For the proposed RH-ISCRC, we set ${\lambda _1} = 0.001$, ${\lambda _2} = 0.001$, $\lambda  = 2.5/{n_a}$ ($n_a$ is the number of samples in the query set), $\gamma=\lambda/2$.
The number of atoms in the compressed set ${\bm{D}}_k$ is set as 20 on Honda/UCSD and 10 on CMU MoBo and YouTube.
For KCH-ISCRC, $\tau=1$ and the number of atoms in each ${\bm{D}}_k$ is set as 50 for all datasets.
The sensitivity of the proposed methods to parameters will be discussed in Section \ref{sensitivity}.

\subsection{Honda/UCSD}
\begin{figure}[h]
\centering
\includegraphics[width=7cm, height=6cm]{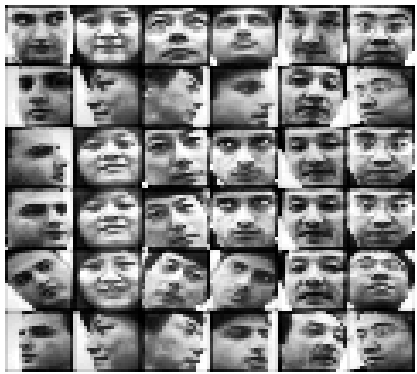}
\caption{Some examples of Honda/UCSD dataset}
\label{Hondaface}
\end{figure}

The Honda/UCSD dataset consists of 59 video sequences involving 20 different subjects \cite{lee2003video}.
The Viola-Jones face detector \cite{viola2004robust} is used to detect the faces in each frame and resize the detected faces to 20$\times$20 images.
Some examples of Honda/UCSD dataset are shown in Figure \ref{Hondaface}.
Histogram equalization is utilized to reduce the illumination variations. Our experiment setting is the same as \cite{lee2003video}\cite{hu2011sparse}: 20 sequences are set aside for training and the remaining 39 sequences for testing. The intensity is used as the feature.

\begin{table}[h]
\begin{center}
\caption {Recognition rates on Honda/UCSD ($\%$)}
\label{honda}
\tabcolsep=0.1in
\begin{tabular}{ccccc}
\hline
Non-kernel	&	50	&	100	&200&Year	\\
\hline
MSM \cite{yamaguchi1998face}	&	74.36	&	79.49&	89.74 &1998\\
DCC \cite{kim2007discriminative} 	&	76.92	&	84.62&94.87&	2007\\
MMD \cite{wang2008manifold} 	&	69.23	&	87.18&94.87&2008\\
MDA \cite{wang2009manifold} 	&	82.05	&	94.87&97.44	&2009\\
SRC \cite{wright2009robust}	&	84.62	&	92.31	&92.31&2009\\
AHISD \cite{cevikalp2010face}	&	82.05	&	84.62&	89.74&2010\\
CHISD \cite{cevikalp2010face}	&	82.05	&	84.62&	92.31&2010\\
SANP \cite{hu2011sparse}	&	84.62	&	92.31&	94.87&2011\\
CRC \cite{zhang2011sparse}	&	84.62	&	94.87&	94.87&2011\\
RNP \cite{yang2013face}	&	87.18	&	94.87&	\textbf{100.0}&2011\\
RH-ISCRC-$l_1$ & \textbf{89.74} &\textbf{97.44}&\textbf{100.0}\\
RH-ISCRC-$l_2$ & \textbf{89.74} &\textbf{97.44}&\textbf{100.0}\\
\hline
Kernel	&	50	&	100&	200&Year	\\
\hline
AHISD \cite{cevikalp2010face}	&	84.62	&	84.62&	82.05&2010\\
CHISD \cite{cevikalp2010face}	&	84.62	&	87.18&	89.74&2010\\
KSRC \cite{gao2010kernel}	&	87.18	&	\textbf{97.44}&	97.44&2009\\
KCRC \cite{yang2013}	&	82.05	&	94.87&	94.87&2012\\
KCH-ISCRC	&	\textbf{89.74}	&	94.87&\textbf{100.0}	&\\
\hline
\end{tabular}
\end{center}
\end{table}

The experimental results are listed in Table \ref{honda}. We can see that for those non-kernel methods, the proposed RH-ISCRC outperforms much all the other methods. For the kernel based method, the proposed KCH-ISCRC performs the best except for the case when 100 frames per set are used. We can also see that on this dataset, RH-ISCRC-$l_1$ and RH-ISCRC-$l_2$ achieve the same recognition rate, which implies that on this dataset the $l_2$-norm regularization is strong enough to yield a good solution to the regularized hull based CRSSD in Eq. (\ref{slc3}).

\subsection{CMU MoBo}
The CMU Mobo\footnote{http://www.ri.cmu.edu/publication$\_$view.html?pub$\_$id=3904} (Motion of Body) dataset \cite{gross2001cmu} was originally established for human pose identification and it contains 96 sequences from 24 subjects. Four video sequences are collected per subject, each of which corresponds to a walking pattern. Again, the Viola-Jones face detector \cite{viola2004robust} is used to detect the faces and the detected face images are resized to 40 $\times$ 40. The LBP feature is used, which is the same as the work in \cite{cevikalp2010face} and \cite{hu2011sparse}.

One video sequence per subject is selected for training while the rest are used for testing. Ten-fold cross validation experiments are conducted and the average recognition results are shown in Table \ref{mobo}. We can clearly see that the proposed methods outperform the other methods under different frames per set. On this dataset and the Honda/UCSD dataset, the proposed non-kernel RH-ISCRC and the kernel based KCH-ISCRC have similar ISFR rates.

\begin{table}[h]
\begin{center}
\caption {Recognition rates on CMU MoBo($\%$)}
\label{mobo}
\tabcolsep=0.05in
\begin{tabular}{ccccccc}
\hline
Non-kernel	&	50	&	100&	200&Year\\
\hline
MSM \cite{yamaguchi1998face}	&	84.3 $\pm$ 2.6	&	86.6$\pm$2.2&89.9$\pm$2.4	&1998\\
DCC \cite{kim2007discriminative} 	&	82.1$\pm$ 2.7	&	85.5$\pm$2.8&91.6$\pm$2.5	&2007\\
MMD \cite{wang2008manifold} 	&	86.2 $\pm$2.9 	&	94.6$\pm$1.9&	96.4$\pm$0.7&2008\\
MDA \cite{wang2009manifold} 	&	86.2 $\pm$2.9 	&	93.2$\pm$2.8&	95.8$\pm$2.3&2009\\
SRC \cite{wright2009robust}	&	91.0 $\pm$2.1 	&	91.8$\pm$2.7&	96.5$\pm$2.5&2009\\
AHISD \cite{cevikalp2010face}	&	91.6 $\pm$2.8 	&	94.1$\pm$2.0&	91.9$\pm$2.6&2010\\
CHISD \cite{cevikalp2010face}	&	91.2 $\pm$3.1 	&	93.8$\pm$2.5&	96.0$\pm$1.3&2010\\
SANP \cite{hu2011sparse}	&	91.9 $\pm$2.7 	&	94.2$\pm$2.1&	97.3$\pm$1.3&2011\\
CRC \cite{zhang2011sparse}	&	89.6 $\pm$1.8 	&	92.4$\pm$3.7&	96.4$\pm$2.8 &2011\\
RNP \cite{yang2013face}	&	91.9 $\pm$2.5 	&	94.7$\pm$1.2&	97.4$\pm$1.5 &2013\\
RH-ISCRC-$l_1$ 	&	\textbf{93.5$\pm$2.8}	&	\textbf{96.5$\pm$1.9}& \textbf{98.7$\pm$1.7}	&\\
RH-ISCRC-$l_2$  &	\textbf{93.5$\pm$2.8}	&	\textbf{96.4$\pm$1.9}& \textbf{98.4$\pm$1.7}	&\\
\hline
Kernel	&	50	&	100	&200	&	Year\\
\hline
AHISD \cite{cevikalp2010face}	&	88.9$\pm$1.7	&	92.4$\pm$2.8& 93.5$\pm$4.2	&2010\\
CHISD \cite{cevikalp2010face}	&	91.5$\pm$2.0	&	93.4$\pm$4.0& 97.4$\pm$1.9	&2010\\
KSRC \cite{gao2010kernel}	&	91.6 $\pm$2.8 	&	94.1$\pm$2.0& 96.8$\pm$2.0	&2010\\
KCRC \cite{yang2013}	&	91.2 $\pm$3.1 	&	93.4$\pm$2.9& 96.6$\pm$2.6	&2012\\
KCH-ISCRC	&	\textbf{94.2 $\pm$2.1} 	&	\textbf{96.4$\pm$2.3}&	\textbf{98.4$\pm$1.9}&\\
\hline
\end{tabular}
\end{center}
\end{table}

\subsection{YouTube Celebrities}
\begin{figure}[h]
\centering
\includegraphics[width=8cm]{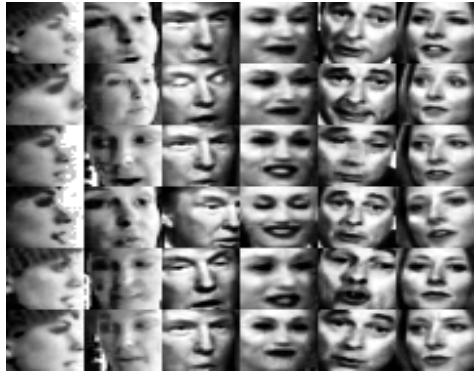}
\caption{Some examples of YouTube Celebrities dataset}
\label{youtubeface}
\end{figure}

The YouTube Celebrities\footnote{http://seqam.rutgers.edu/site/index.php?option=com$\_$content$\&$view=article$\&$id =64$\&$Itemid=80} is a large scale video dataset collected for face tracking and recognition, consisting of 1,910 video sequences of 47 celebrities from YouTube \cite{kim2008face}.
As the videos were captured in unconstrained environments, the recognition task becomes much more challenging due to the larger variations in pose, illumination and expressions.
Some examples of YouTube Celebrities dataset are shown in Figure \ref{youtubeface}.
The face in each frame is also detected by the Viola-Jones face detector and resized to a 30 $\times$ 30 gray-scale image.
The intensity value is used as feature.
The experiment setting is the same as \cite{hu2011sparse,wang2009manifold,wang2012covariance}.
Three video sequences per subject are selected for training and six for testing.
Five-fold cross validation experiments are conducted.

The experimental results are shown in Table \ref{youtube1}.
It can be seen that among the non-kernel methods, the proposed RH-ISCRC-$l_1$ achieves the highest recognition rate, while among the kernel based methods, the proposed KCH-ISCRC performs the best.
Since this Youtube Celebrities dataset was established under uncontrolled environment, there are significant variations among the query and gallery sets, and therefore the $l_1$-regularization is very helpful to improve the stability and discrimination of the solution to Eq. (\ref{slc3}).
As a consequence, RH-ISCRC-$l_1$ leads to much better results than RH-ISCRC-$l_2$ on this dataset.
On the other hand, the kernel based KCH-ISCRC leads to better results than RH-ISCRC in this experiment.
Besides, the number of frames per set also affect the performance of ISCRC.
When number of frames is small, the improvement by ISCRC is more significant.
\begin{table}[h]
\begin{center}
\caption {Recognition rates on YouTube (V1 $\%$)}
\label{youtube1}
\tabcolsep=0.05in
\begin{tabular}{ccccccc}
\hline
Non-kernel	&	50	&	100&	200&Year\\
\hline
MSM \cite{yamaguchi1998face}	&	54.8$\pm$8.7	&	57.4$\pm$7.7	&	56.7$\pm$6.9	&	1998	\\
DCC \cite{kim2007discriminative}	&	57.6$\pm$8.0	&	62.7$\pm$6.8	&	65.7$\pm$7.0	&	2007	\\
MMD \cite{wang2008manifold} 	&	57.8$\pm$6.6	&	62.8$\pm$6.2	&	64.7$\pm$6.3	&	2008	\\
SRC	\cite{wright2009robust} &	61.5$\pm$6.9	&	64.4$\pm$6.8	&	66.0$\pm$6.7	&	2009	\\
MDA \cite{wang2009manifold}	&	58.5$\pm$6.2	&	63.3$\pm$6.1	&	65.4$\pm$6.6	&	2009	\\
AHISD \cite{cevikalp2010face}	&	57.5$\pm$7.9	&	59.7$\pm$7.2	&	57.0$\pm$5.5	&	2010	\\
CHISD \cite{cevikalp2010face}	&	58.0$\pm$8.2	&	62.8$\pm$8.1	&	64.8$\pm$7.1	&	2010	\\
SANP \cite{hu2011sparse}&	57.8$\pm$7.2	&	63.1$\pm$8.0	&	65.6$\pm$7.9	&	2011	\\
CRC	\cite{zhang2011sparse}&	56.5$\pm$7.4	&	59.5$\pm$6.6	&	61.4$\pm$6.4	&	2011	\\
RNP \cite{yang2013face}	&	59.9 $\pm$7.3 	&	63.3$\pm$8.1&	64.4$\pm$7.8 &2013\\
RH-ISCRC-$l_1$	&	\textbf{62.3$\pm$6.2}	&	 \textbf{65.6$\pm$6.7}	&	\textbf{66.7$\pm$6.4}	&		\\
RH-ISCRC-$l_2$	&	57.4$\pm$7.2	&	 60.7$\pm$6.5	&	61.4$\pm$6.4	&		\\
\hline
Kernel	&	50	&	100&	200&Year	\\
\hline
AHISD	\cite{cevikalp2010face}&	57.2$\pm$7.5	&	59.6$\pm$7.4	&	61.8$\pm$7.3	&	2010	\\
CHISD	\cite{cevikalp2010face}&	57.9$\pm$8.3	&	62.6$\pm$8.1	&	64.9$\pm$7.2	&	2010	\\
KSRC	\cite{gao2010kernel}&	61.4$\pm$7.0	&	65.9$\pm$6.9	&	67.8$\pm$6.4	&	2010	\\
KCRC	\cite{yang2013}&	57.5$\pm$7.9	&	60.6$\pm$6.8	&	62.7$\pm$7.7	&	2012	\\
KCH-ISCRC	&	\textbf{64.5$\pm$7.6}	&	\textbf{67.4$\pm$8.0}	&	\textbf{69.7$\pm$7.4}	&		\\
\hline
\end{tabular}
\end{center}
\end{table}

\subsection{Time comparison}
Then let's compare the efficiency of competing methods.
The Matlab codes of all competing methods are obtained from the original authors, and we run them on an Intel(R) Core(TM) i7-2600K (3.4GHz) PC.
The average running time per set on CMU MoBo (200 frames per set) is listed in Table \ref{time}.
We can see that the proposed RH-ISCRC-$l_2$ is the fastest among all competing methods except for RNP, while RH-ISCRC-$l_1$ also has a fast speed.
Among all the kernel based methods, the proposed KCH-ISCRC is much faster than others.
Overall, the proposed RH-ISCRC and KCH-ISCRC methods have not only high ISFR accuracy but also high efficiency than the competing methods.
\begin{table}[h]
\begin{center}
\caption {Average running time per set on CMU MoBo ($s$)}
\label{time}
\tabcolsep=0.05in
\begin{tabular}{ccccccc}
\hline
Non-kernel	&	Time & Kernel	&	Time \\
\hline
MSM \cite{yamaguchi1998face}	&	0.338 & AHISD \cite{cevikalp2010face}		&	18.546\\
DCC \cite{kim2007discriminative} 		&	0.349&CHISD \cite{cevikalp2010face}		&		18.166\\
MMD \cite{wang2008manifold} 	&		10.223&KSRC \cite{gao2010kernel}	&		35.508 \\
SRC \cite{wright2009robust}	&		5.301&KCRC \cite{yang2013}	&		6.543\\
MDA \cite{wang2009manifold} 	&	7.031	&KCH-ISCRC	&		\textbf{2.03}\\
AHISD \cite{cevikalp2010face}	&		31.365\\
CHISD \cite{cevikalp2010face}		&		18.029\\
SANP \cite{hu2011sparse}	&		11.124\\
CRC \cite{zhang2011sparse}	&		0.684\\
RNP \cite{yang2013face}	&		\textbf{0.113}\\
RH-ISCRC-$l_1$ &		0.788&\\
RH-ISCRC-$l_2$ &		\textbf{0.280}&\\
\hline
\end{tabular}
\end{center}
\end{table}

\subsection{Parameter sensitivity analysis}
\label{sensitivity}
\begin{figure}[h]
\centering
\includegraphics[width=9cm]{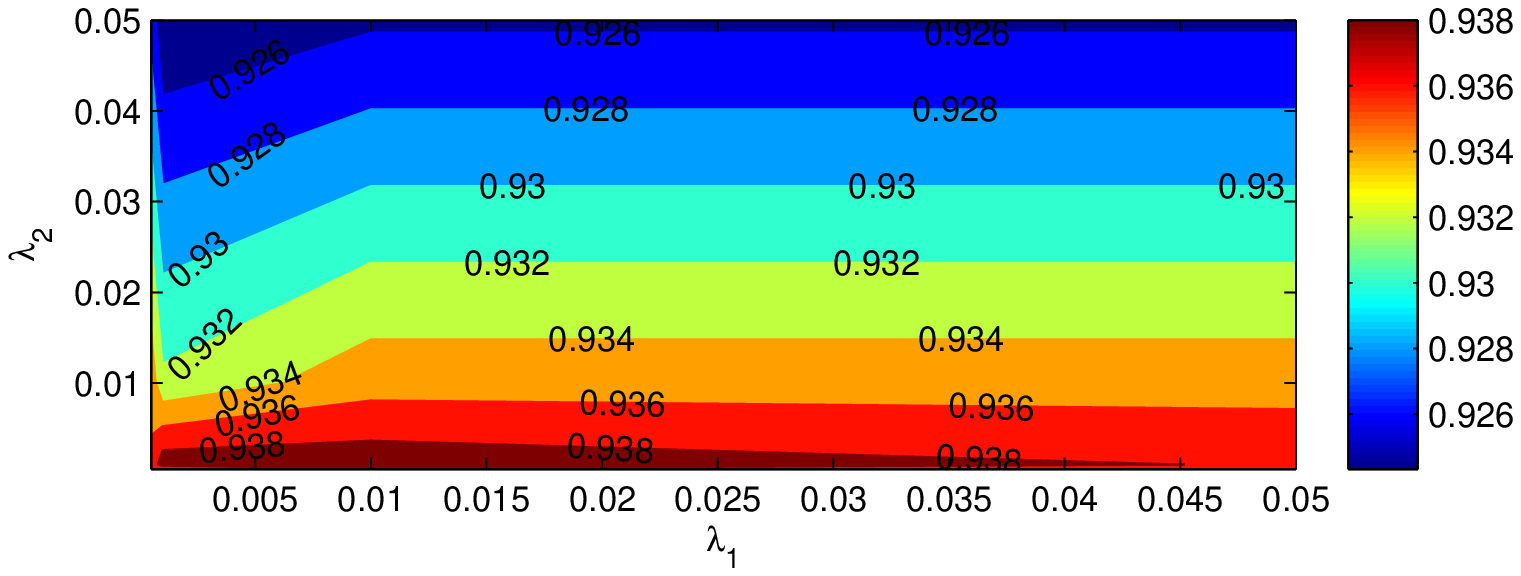}\\
\includegraphics[width=9cm]{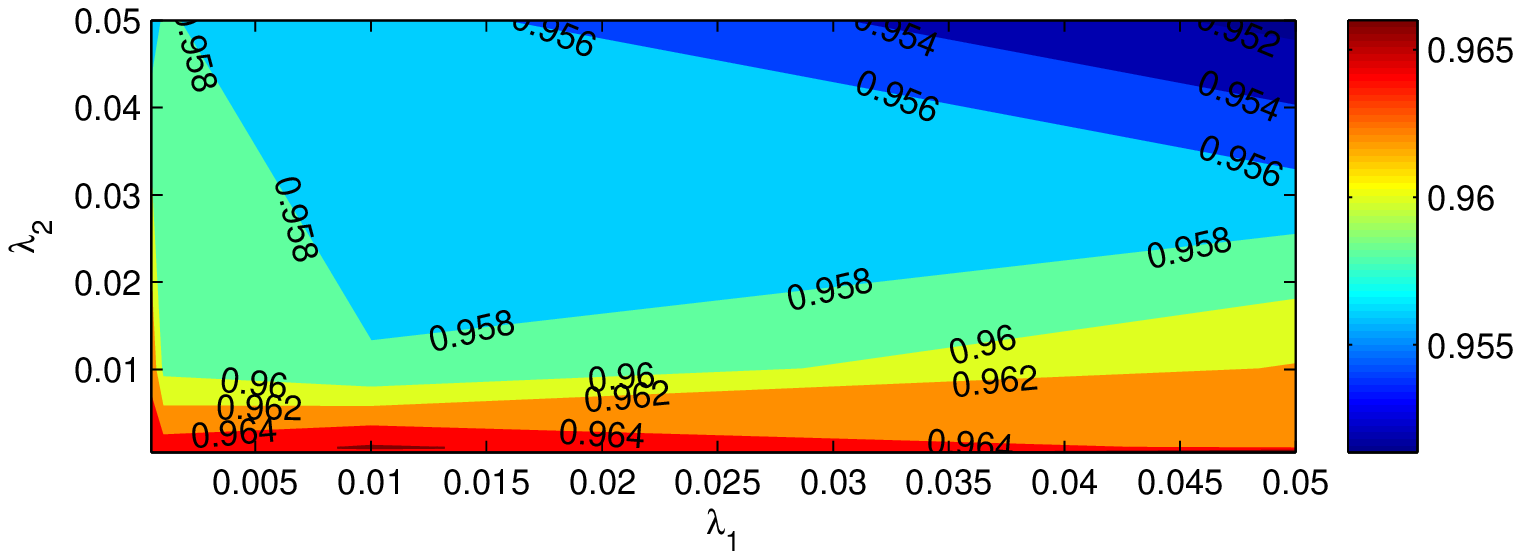}\\
\includegraphics[width=9cm]{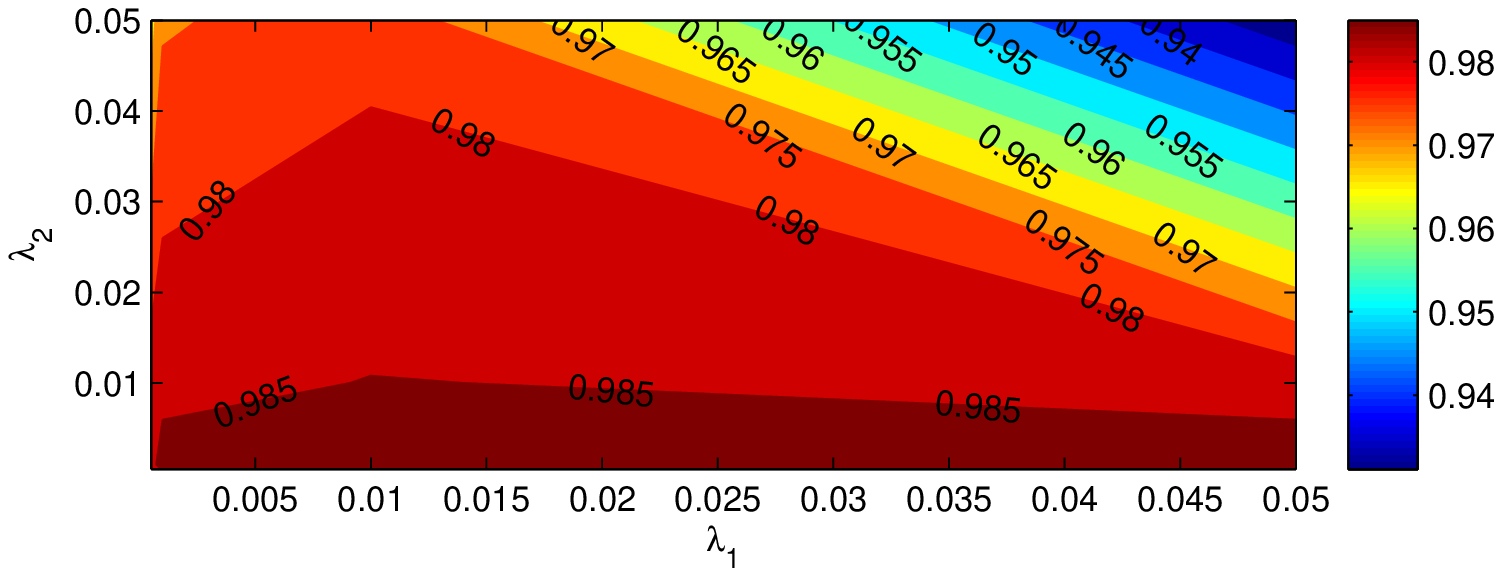}\\
\caption{Recognition performance of RH-ISCRC-$l_1$ on CMU MoBo with different $\lambda_1$ and $\lambda_2$. Different colors represent different accuracy. Top: 50 frames per set; middle: 100 frames per set; bottom: 200 frames per set.}
\label{time1}
\end{figure}

\begin{figure}[H]
\centering
\includegraphics[width=9cm]{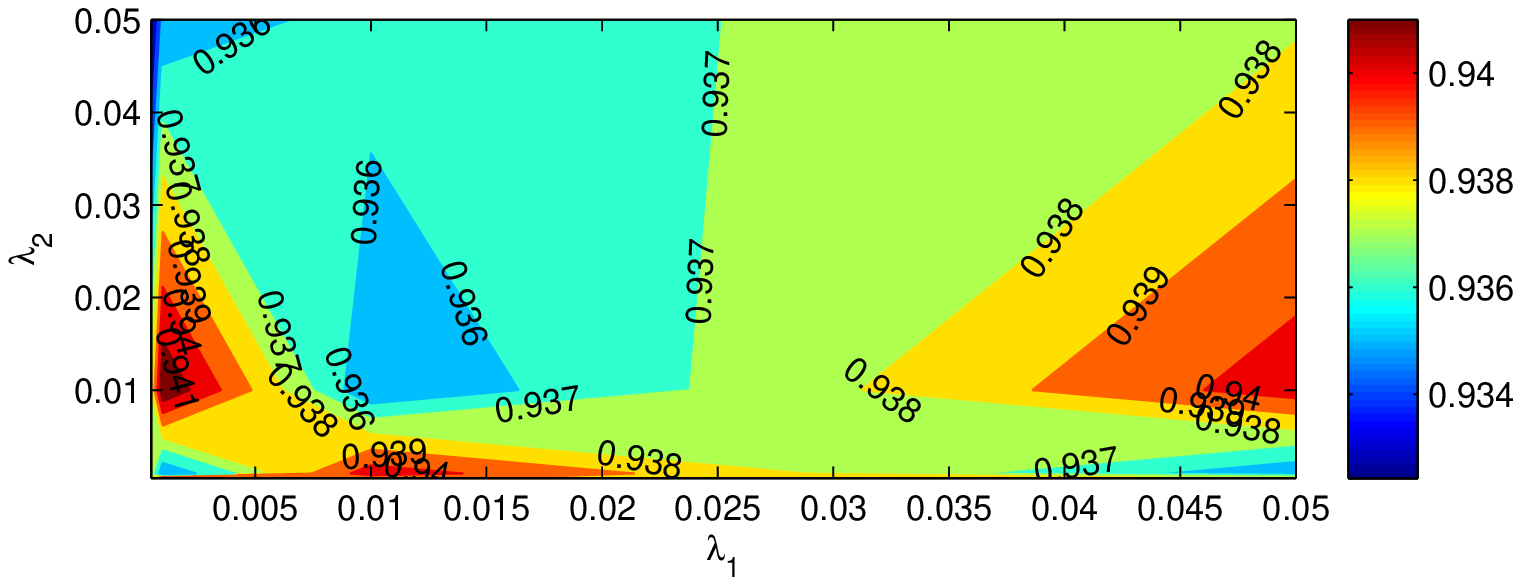}\\
\includegraphics[width=9cm]{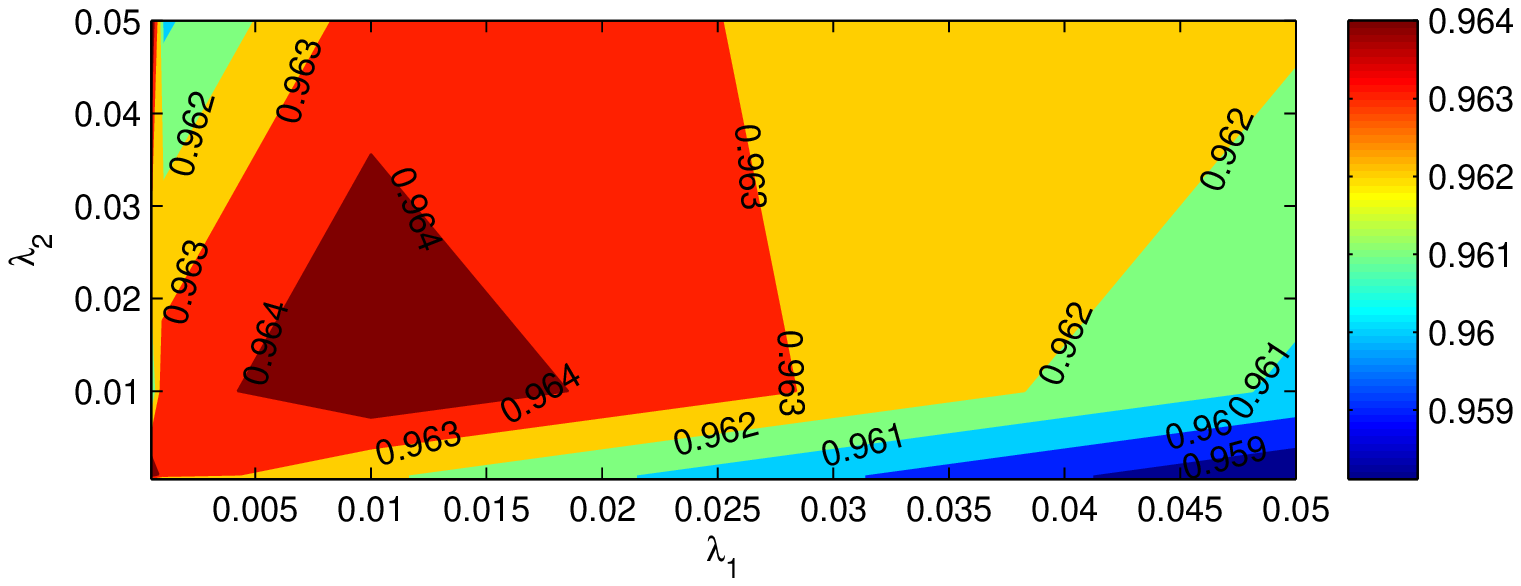}\\
\includegraphics[width=9cm]{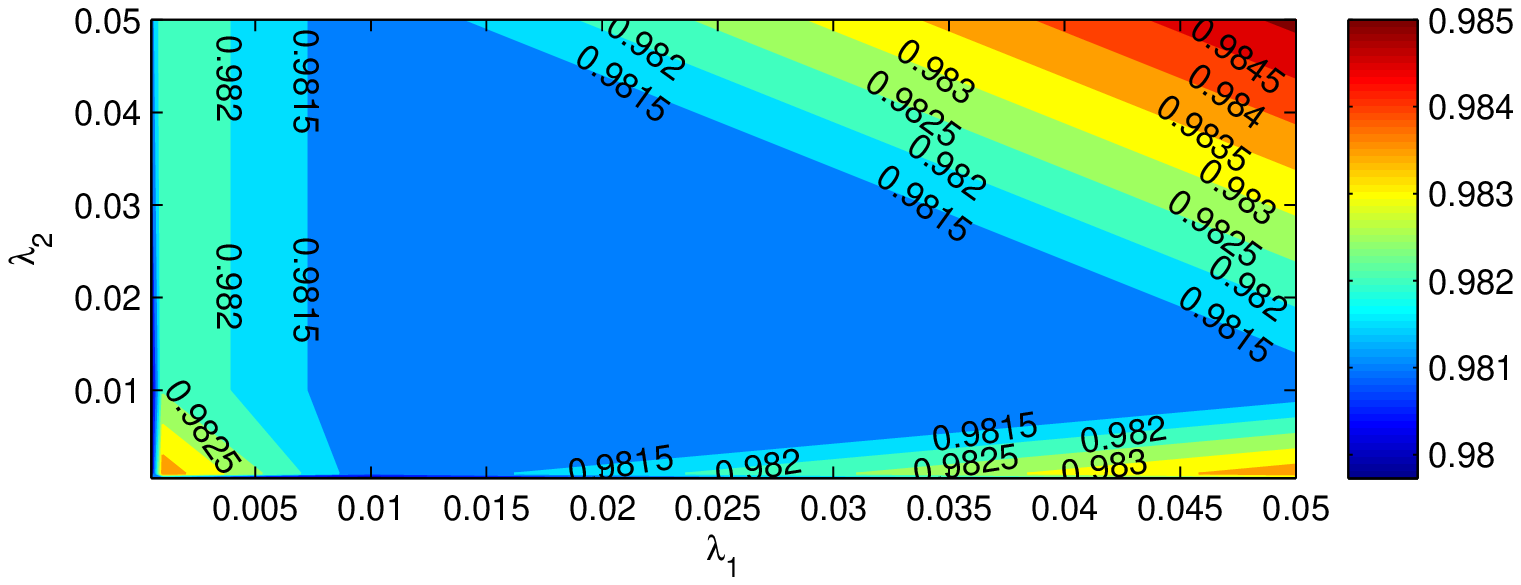}\\
\caption{Recognition performance of RH-ISCRC-$l_2$ on CMU MoBo with different $\lambda_1$ and $\lambda_2$. Different colors represent different accuracy. Top: 50 frames per set; middle: 100 frames per set; bottom: 200 frames per set.}
\label{time2}
\end{figure}

\begin{figure}[H]
\centering
\includegraphics[width=8cm]{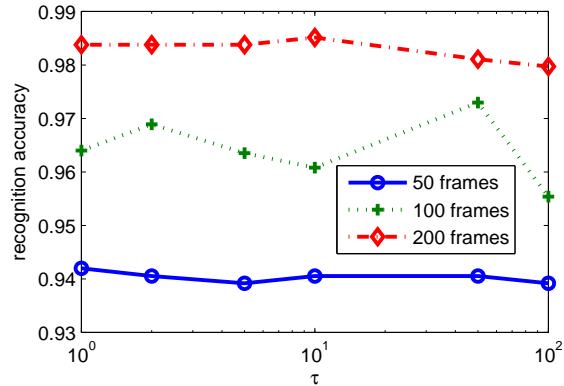}
\caption{Recognition performance of KCH-ISCRC on CMU MoBo with different $\tau$.}
\label{time3}
\end{figure}

To verify if the proposed methods are sensitive to parameters, in this section we present the recognition accuracies with different parameter values.
For RH-ISCRC, there are two parameters, ${\lambda _1}$ and ${\lambda _2}$ in Eq. (\ref{admover}), which need to be set.
For KCH-ISCRC, there is only one parameter $\tau$ in Eq. (\ref{CHC}).
We show the recognition accuracies versus the parameters on the CMU MoBo dataset in Fig. \ref{time1}, Fig. \ref{time2} and Fig. \ref{time3}, respectively, for RH-ISCRC-$l_1$, RH-ISCRC-$l_2$ and KCH-ISCRC.
The different colors correspond to different accuracies, as shown in the color bar.
${\lambda _1}$ and ${\lambda _2}$  are selected from $\{0.0005, 0.001,0.01,0.05\}$.
In Fig. \ref{time1} and Fig. \ref{time2}, the top sub-figure is for 50 frames per set, the middle is for 100 frames per set and the bottom corresponds to 200 frames per set.
From Fig. \ref{time1}, we can see that the accuracy of RH-ISCRC-$l_1$ is very stable when ${\lambda _1}$ varies from 0.0005 to 0.05 and ${\lambda _2}$ varies from 0.0005 to 0.01.
When ${\lambda _2}$ is increased to 0.05, the recognition performance would degrade.
Fig. \ref{time2} shows that RH-ISCRC-$l_2$ is insensitive to the values of ${\lambda _1}$ and ${\lambda _2}$.
For example, in the experiments of 100 and 200 frames per set, the accuracy variation is within 0.5$\%$ for different $\lambda_1$ and $\lambda_2$.
Considering the performance of both RH-ISCRC-$l_1$ and RH-ISCRC-$l_2$, ${\lambda _1}$ and ${\lambda _2}$ can both be set as 0.001.
With this parameter setting, the accuracy is very stale in different experiments.
For KCH-ISCRC, its recognition accuracies with different values of $\tau$ are shown in Fig. \ref{time3}.
$\tau$ is set as $\{1, 2, 5, 10, 50, 100\}$.
One can see that KCH-ISCRC is insensitive to $\tau$.
Hence, we simplely set $\tau$ as 1.

\section{Conclusion}
\label{section6}
We proposed a novel image set based collaborative representation and classification (ISCRC) scheme for image set based face recognition (ISFR). The query set was modeled as a convex or regularized hull, and a collaborative representation based set to sets distance (CRSSD) was defined by representing the hull of query set over all the gallery sets. The CRSSD considers the correlation and distinction of sample images within the query set and the relationship between the gallery sets. With CRSSD, the representation residual of the hull of query set by each gallery set can be computed and used for classification. Experiments on the three benchmark ISFR databases showed that the proposed ISCRC is superior to state-of-the-art ISFR methods in terms of both recognition rates and efficiency.

\section*{Acknowledgment}
The authors thank T. Kim for sharing the source code of DCC, S. Gao for the source code of KSRC and Y. Hu for the source code of SANP.
We thank R. Wang for sharing the source code of MMD and MDA, and the cropped faces of the Honda/UCSD dataset and YouTube Celebrities dataset.
We also thank H. Cevikalp for sharing the source code of AHISD/CHISD and providing the LBP features for the Mobo dataset.

\ifCLASSOPTIONcaptionsoff
  \newpage
\fi

\bibliographystyle{IEEEtran}
\bibliography{egbib,reference}

\end{document}